# The "Criminality from Face" Illusion

Kevin W. Bowyer, *Fellow, IEEE*, Michael C. King, *Member, IEEE,* and Walter Scheirer, *Senior Member, IEEE, Kushal Vangara*

*Abstract*— The automatic analysis of face images can generate predictions about a person's gender, age, race, facial expression, body mass index, and various other indices and conditions. A few recent publications have claimed success in analyzing an image of a person's face in order to predict the person's status as Criminal / Non-Criminal. Predicting "criminality from face" may initially seem similar to other facial analytics, but we argue that attempts to create a criminality-from-face algorithm are necessarily doomed to fail, that apparently promising experimental results in recent publications are an illusion resulting from inadequate experimental design, and that there is potentially a large social cost to belief in the criminality from face illusion.

## I. INTRODUCTION

CRIMINAL or not? Is it possible to create an algorithm that analyzes an image of a person's face and accurately labels the person as Criminal or Non-Criminal? Recent research tackling this problem has reported accuracy as high as 97% [14] using convolutional neural networks (CNNs). In this paper, we explain why the concept of an algorithm to compute "criminality from face," and the high accuracies reported in recent publications, are an illusion.

Facial analytics seek to infer something about an individual other than their identity. Facial analytics can predict, with some reasonable accuracy, things such as age [10], gender [6], race [9], facial expression / emotion [25], body mass index [5], and certain types of health conditions [29]. A few recent papers have attempted to extend facial analytics to infer criminality from face, where the task is to take a face image as input, and predict the status of the person as Criminal / Non-Criminal for output. This concept is illustrated in Figure 1.

One of these papers states that "As expected, the state-of-the-art CNN classifier performs the best, achieving 89.51% accuracy...These highly consistent results are evidences for the validity of automated face-induced inference on criminality, despite the historical controversy surrounding the topic" [40]. Another paper states that, "the test accuracy of 97%, achieved by CNN, exceeds our expectations and is a clear indicator of the possibility to differentiate between criminals and non-criminals using their facial images" [14]. (During the review period of this paper, we were informed by one of the authors of [14] that they had agreed with the journal to retract their paper.) A press release about another paper titled "A Deep Neural Network Model to Predict Criminality Using Image Processing" stated that "With 80 percent accuracy and with no racial bias, the software can predict if someone is a criminal based solely on a picture of their face. The software is intended to help law enforcement prevent crime." The original press release generated so much controversy that it "was removed from the website at the request of the faculty involved" and replaced by a statement meant to defuse the situation: "The faculty are updating the paper to address concerns raised" [13].

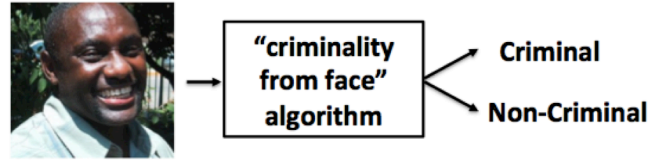

Fig. 1. The "Criminality from Face" Concept. The concept is that there are features that can be extracted by analysis of a facial image and used to categorize the person in the image as Criminal or Non-Criminal. The image on the left is of Thomas Doswell, who served 19 years in jail for a crime he did not commit, and was freed based on DNA evidence. See the Innocence Project [50] for additional details.

Section II of this paper explains why the concept of an algorithm to compute criminality from face is an illusion. A useful solution to any general version of the problem is impossible. Sections III and IV explain how the impressive reported accuracy levels are readily accounted for by inadequate experimental design that has extraneous factors confounded with the Criminal / Non-Criminal labeling of images. Learning incidental properties of datasets rather than the intended concept is a well-known problem in computer vision. Section V explains how Psychology research on first impressions of a face image has been mis-interpreted as suggesting that it is possible to accurately characterize true qualities of a person. Section VI briefly discusses the legacy of the Positivist School of criminology. Lastly, Section VII describes why the belief in the illusion of a criminality-from-face algorithm potentially has large, negative consequences for society.

## II. AN ILLUSORY PROBLEM DEFINITION

Part of the criminality from face illusion is that the problem definition is simple to state and is similar in form to that of facial analytics with sound foundations. However, simple thought experiments reveal the impossibility of creating any general algorithm to correctly apply the label Criminal / Non-Criminal to a face.

Consider a person who to a given point in their life has never even thought of committing a crime. Assume Image A is their face from this period. (See Figure 2.) One day this person is inspired by what they imagine to be the perfect crime. From the moment that idea enters their head and heart, they know that they will commit the crime. Image B is their face from this period. The fateful day arrives, the crime is committed, and life proceeds well for a while. Image C is their face from this period. But the day of arrest, trial and conviction comes, and the person begins to serve their sentence. Image D is their face from this period. Eventually they serve their sentence and are released. Image E is from this period. The criminality-from-face algorithm developer now confronts the question: what is the ground-truth label for each of the images?

Answering this question forces the algorithm developer to face up to (pun intended) the fact that there is no plausible foundation for the criminality-from-face problem definition. One possible answer is to assign all four images the label of Criminal. This requires the algorithm developer to believe in the "born criminal" concept and also to believe that there is something measurable from the face image that reveals this predestined condition of a person. There have been historical criminologists who subscribed to these beliefs; for example, Cesare Lombroso, who will be discussed later. But today these beliefs are regarded as having no scientific foundation [47]. Another possible answer is that Image A should be labeled Non-Criminal and B to E should be labeled Criminal. This answer requires the algorithm developer to believe that the criminal intent entering the person's head and heart causes a measurable change in their facial appearance. Still another possible answer is that Images A and B should be labeled Non-Criminal and C to E labeled Criminal. This answer requires the algorithm developer to believe that the act of committing a crime causes a measurable change in the person's facial appearance. Another answer assigns Images A to C to Non-Criminal and Images D and E to Criminal. This answer requires the algorithm developer to believe that being convicted of the crime causes a measurable change in the person's facial appearance. And if conviction causes a change, then does completing a sentence and being released reverse that change in Image E? These last possible answers are so untenable that we do not know of anyone who advocates for any of them. However, believing that a criminality-from-face algorithm can exist requires believing there is a rational assignment of labels to the images.

A second thought experiment highlights additional difficulties in the criminality-from-face problem definition. In the United States, marijuana use is illegal under federal law. However, about fifteen states have currently decriminalized recreational use of marijuana, and a majority of states have laws that allow medical use of marijuana. The trend is toward additional states decriminalizing recreational use and allowing medical use. Should a criminality-from-face algorithm operate according to federal law or according to state law? If the algorithm is to operate according to state law, what should the algorithm do with images taken before and after a change in state law? The problem highlighted here is that Criminality / Non-Criminality can be a social construct that can vary with the location and over time. Instances involving more serious crimes than drug use include the killing of a person and the use of "stand your ground" laws in determining criminality, and the process of marital rape becoming a crime in the various states of the US between the mid-1970s and mid-1990s.

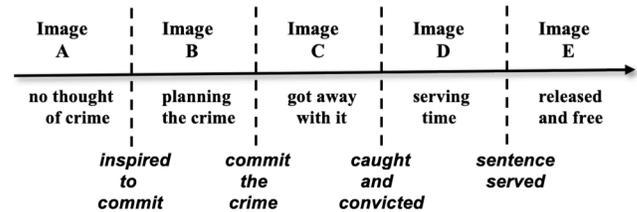

Fig. 2. Images from different points in a person's life. The "criminality from face" algorithm developer must decide on ground-truth labels for the images, which reveals the algorithm developer's belief about the cause of the relevant features.

Reflecting on these examples, there is nothing about conceiving, committing, or being convicted of a crime that causes any distinctive change in facial appearance. There is no distinctive feature of facial appearance that predestines a person to become a criminal or to be unable to become a criminal. Whether or not a given action is sufficient to allow a person to be convicted of a crime can depend on the time and location where the action is committed. What, then, should be made of reports that algorithms have achieved impressive levels of accuracy in labeling images as Criminal / Non-Criminal? Nothing more than a basic flaw in experimental method is required to explain the apparently impressive accuracy results.

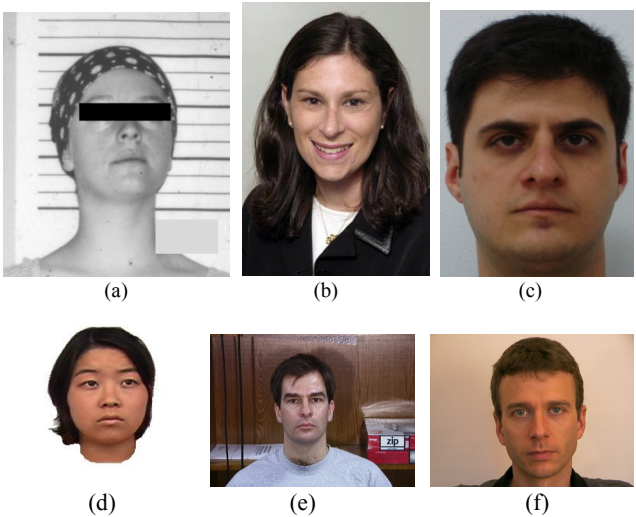

Fig. 3. Example Images: (a) From a NIST mugshot dataset, labeled "Criminal" in [14]; (b) to (f), from FDDB, FEI, Face Place, GT and MIT CBCL datasets, respectively, labeled "Non-Criminal" in [14]. Black rectangle added to eye region of original image (a) in respect of the privacy of the individual.

## III. ILLUSORY EXPERIMENTAL RESULTS

The reasoning that criminality-from-face is just an extension of existing sound facial analytics leads to flawed experimental designs. In predicting age from a facial image, there are known, identifiable features that correlate with increased age (e.g., lines on the face), and a person's age is the same regardless of their geographic location. Similarly, in predicting a person's emotional state from a facial image, there are known configurations of facial features that correspond to particular emotions, and no expectation that an emotion detected in one image reveals a permanent condition of a person's life. Criminality-from-face algorithms do not share any of the firm foundational elements of other, sound facial analytics.

To better understand what goes wrong in these cases, let's begin with a close look at the data and experiments for the criminality-from-face algorithm from the paper by Hashemi and Hall [14]. This paper is important to discuss because: (1) the authors recognize the potential for controversy, as they state "...this study's scope is limited to the technical and analytical aspects of this topic, while its social implications require more scrutiny and its practical applications demand even higher levels of caution and suspicion"; (2) remarkably high accuracy is reported, with the authors declaring that "the test accuracy of 97%, achieved by CNN, exceeds our expectations and is a clear indicator of the possibility to differentiate between criminals and non-criminals using their facial images"; and (3) the paper appears in a peer-reviewed journal owned by a well-regarded publisher, so we can assume that the reviewers and editor, as well as the authors, believed in the validity of the work.

Experimental work on criminality-from-face algorithms naturally requires a dataset of face images, some of which are labeled Criminal and some Non-Criminal. An algorithm is trained on a subset of this data and the accuracy of the resulting algorithm should be estimated on a different subset of the data. It is of course essential to avoid possible sources of bias in the data. There should be no extraneous differences between images in the Criminal and the Non-Criminal categories. As a trivial example, if all persons in images labeled criminal wore black hats and all persons in images labeled Non-Criminal wore white hats, the algorithm might learn the difference in color of hats, and 100% accuracy might be reported, when in fact the algorithm is useless at detecting criminals. Meticulous attention to the details of the experimental dataset is essential, even more so when training deep neural networks than it is when using "hand-crafted" features. Deep neural networks will by their nature pick up on any consistent difference between images in the two categories.

The face images for the Criminal category are described as follows [14]. "A total of 8401 grayscale mugshot images of arrested individuals are obtained from National Institute of Standards and Technology (NIST) Special Database. Images are all in png format ... Cropping the facial rectangle from the rest of the image prevents the classifier from being affected by peripheral or background effects surrounding the face...The result contains 5000 front view face images of 4796 male and 204 female individuals and of variable sizes, ranging from 238 × 238 up to 813 × 813 pixels. Since neural networks receive inputs of the same size, all images are resized to 128 × 128."

The web page for NIST Special Database 18 [19] states that it contains "... 3248 segmented 8-bit gray scale mugshot images (varying sizes) of 1573 individuals". The source of the discrepancy in number of persons and images in the Criminal category in [14] and in the NIST dataset [19] is not known to us. An additional detail is that the User's Guide for NIST Special Database 18 [38] states that a Kodak MegaPixel1 camera was used to digitize printed mugshot photos.

The face images for the Non-Criminal category are described as follows [14]. "A total of 39,713 RGB facial images are obtained from five sources (Face Recognition Database, FEI Face Database, Georgia Tech face database, Face Place, Face Detection Data Set and Benchmark) ... The images are then converted to grayscale, again to be compatible with mugshots in the criminal dataset. The result contains 5000 front view face images of 3727 male and 1273 female individuals and of variable size, ranging from 87 × 87 up to 799 × 799 pixels. Images are resized to 128 × 128."

The images in the criminal and the Non-Criminal category are all size 128×128, all grayscale, all nominally frontal pose and neutral expression. These factors may make it seem that differences between the images in the two categories are controlled. However, based on the descriptions of the data, there are also multiple extraneous factors that have 100% correlation with the two categories of images.

- All images for the Criminal category come from the NIST dataset, and all images for the Non-Criminal category come from a set of five datasets from other sources.

- All of the images labeled Criminal are photographs of printed images and are taken in a controlled manner with the same camera model, and all of the images labeled Non-Criminal are photographs of live persons taken by various cameras.
- All of the images labeled Criminal were in (lossless) PNG format, and all of the images labeled Non-Criminal were in (lossy) JPG format.
- All of the images labeled criminal started out as grayscale; all of the images labeled Non-Criminal were converted from color to grayscale by the investigators.

Rather than the CNN learning to distinguish between Criminal and Non-Criminal faces, it could have learned to distinguish between (a) images converted to grayscale using the tool the investigators used, and grayscale images from some other source, (b) images originally in PNG and images originally in JPG, (c) images of printed pictures of persons versus images of live persons, or some other property also completely unrelated to the Criminal / Non-Criminal categorization.

Also, as detailed in the User's Guide for the NIST dataset, the mugshots for all "criminal" face images were initially printed photographs that were digitized using an identical process and camera. There are several studies in automated forensic analysis that exploit unique photo response non-uniformity (PRNU) noise characteristics embedded in images to enable camera identification (i.e., device fingerprinting) [3]. The conventional methods have evolved to include CNN based architectures [4]. Thus, the experiment in [14] may simply show an ability to detect printed mugshot images digitized using the Kodak MegaPixel1 camera.

To be fair, Hashemi and Hall note the existence of confounding factors, before dismissing the possibility that this had a significant effect on the results [14]. "It is noteworthy that the criminal mugshots are coming from a different source than non-criminal face shots. That means the conditions under which the criminal images are taken are different than those of non-criminal images. These different conditions refer to the camera, illumination, angle, distance, background, resolution, etc. Such disparities which are not related to facial structure though negligible in majority of cases, might have slightly contributed in training the classifier and helping the classifier to distinguish between the two categories. Therefore, it would be too ambitious to claim that this accuracy is easily generalizable" (italics added). However, given the number of obvious disparities between the two categories, there is no good reason to believe that the CNN was able to learn a model of Criminal / Non-Criminal facial structure. We believe that it is infinitely more likely that the "disparities not related to facial structure" are the only thing that the CNN is using to separate the two categories of images.

The experimental dataset used by Wu and Zhang [40] has similar problems. The Non-Criminal images for that work are described as follows. "Subset $S_n$ contains ID photos of 1126 non-criminals that are acquired from Internet using the web spider tool; they are from a wide gamut of professions and social status, including waiters, construction workers, taxi and truck drivers, real estate agents, doctors, lawyers and professors; roughly half of the individuals in subset $S_n$ have university degrees." But the Criminal images come from specialized sources. "Subset $S_c$ contains ID photos of 730 criminals, of which 330 are published as wanted suspects by the ministry of public security of China and by the departments of public security for the provinces of Guangdong, Jiangsu, Liaoning, etc.; the others are provided by a city police department in China under a confidentiality agreement. We stress that the criminal face images in $S_c$ are normal ID photos not police mugshots. Out of the 730 criminals 235 committed violent crimes including murder, rape, assault, kidnap and robbery; the remaining 536 are convicted of non- violent crimes, such as theft, fraud, abuse of trust (corruption), forgery and racketeering." The essential point is that if there is anything at all different about ID photos acquired from the Internet versus ID photos supplied by a police department, this difference is 100% correlated with the Criminal / Non- Criminal labels and will be used by the trained CNN to classify the images. So, just as with the experiments in [14], there is no good reason to believe that the CNN in the experiments in [40] was able to learn a model of Criminal / Non-Criminal facial structure.

Beyond the problem of extraneous factors that are 100% correlated with the image categories labeled Criminal and Non-Criminal, there is the problem that the image categories do not in fact have the suggested mapping to the real world. The face images used by Wu and Zhang [40] to represent Criminal are stated to be persons who committed crimes, whereas the images used for Criminal in the experiments in [14] are only mugshot images. In the United States, a mugshot is taken when a person is arrested and arrives at a booking station; the person may or may not have committed a crime of been convicted of a crime. Searching for "criminal," "convicted," or "guilty" in the README file for NIST Special Database 18 [38] yields no hits.

Further, events in many places and times are well documented where people have been arrested while opposing unjust laws and policies. The actions taken in peaceful protest by the likes of Rosa Parks, Martin Luther King Jr., and John Lewis, which led to their arrest in the 1950s and 1960s, are widely celebrated today. While their actions may no longer be considered as criminal, mugshot images of their faces are widely available on the internet. Equating mugshots to criminality is also problematic because roughly 95% of convictions in the US are based on a defendant's acceptance of a plea deal. It is reported that "...15 percent of all exonerees — people convicted of crimes later proved to be innocent — originally pleaded guilty. That share rises to 49 percent for people exonerated of manslaughter and 66 percent for those exonerated of drug crimes" [28]. So even a mugshot of someone convicted of a crime, may in fact be an image of an innocent person. The converse is also true of the Non-Criminal image set, of course. There is no way to verify that every image in the Non-Criminal category represents a person who has never committed a crime.

## IV. Experimental Illustration of Dataset Bias

The concern that unintentional bias in experimental datasets in computer vision can lead to impressive but illusory results is very familiar to researchers. Torralba and Efros explored the pervasive nature of the problem a decade ago in a well-known paper [36]. They showed that standard classifiers could often achieve surprisingly high accuracy at categorizing the dataset an image belongs to. They pose a "fundamental question" that is highly appropriate in the current context [36]: "However, there is a more fundamental question: are the datasets measuring the right thing, that is, the expected performance on some real-world task? Unlike datasets in machine learning, where the dataset is the world, computer vision datasets are supposed to be a representation of the world."

We designed a small two-step experiment to illustrate the point that a classifier can learn to accurately classify face images derived from different datasets, without exploiting characteristics unique to an individual's face. In the first step of this experiment, mugshot images from the NIST SD18 dataset represent one class as in [14], and images from the Labeled Faces in the Wild (LFW) dataset [48] represent the other class (See Figure 4a and 4b).

We use simple features and train a simple support vector machine (SVM) classifier to accurately distinguish between the two classes. We extract 1721 frontal face images from each of the NIST SD18 and LFW datasets. The histogram of oriented gradients (HOG) operator is applied to extract feature representations for each image. The HOG technique counts occurrences of gradient orientation in localized portions of an image. Details of the feature extraction used are as follows:

- image dimension: 128x128
- HOG orientations: 10
- pixels per cell: (8, 8)
- cells per block: (2, 2)
- HOG feature dimension: 9000

A binary SVM classifier was trained on 60% of the images, selected at random, from both SD18 and LFW. LFW images were labeled as class 0 (non-criminal), and SD18 images were labeled as class 1 (criminal) for training. The accuracy was measured using the remaining 40% of the images, so training and test sets are disjoint. The HOG+SVM classifier yielded an accuracy of 95.86%. These results are congruent with those presented in [14], but we hope it is clear that the classifier does not encode any notion of criminality.

As a second step in our experiment, we tested our trained SVM classifier on images from the FERET dataset [49] (Figure 4c). We expect that none, or almost none, of the subjects in the FERET dataset have been convicted of any crimes. We randomly selected 689 (approximately 40% of 1721, so the same size as the test set in the first experiment) frontal images from the FERET dataset, converted them to grayscale, resized to 128x128, and applied the same HOG operator. Applying our trained classifier to this dataset results in 37% of the FERET images being classified as non-criminal (class 0) and 63% of FERET images are classified as criminal (class 1). This result underscores the complete incompetence of the trained classifier when applied to a dataset that it was not trained on. The papers that have advanced the criminality-from-face concept do not conduct this kind of experiment as a sensibility check.

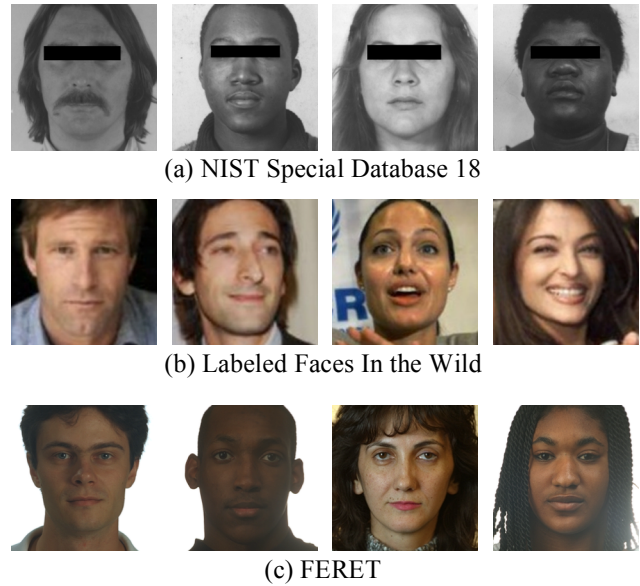

(a) NIST Special Database 18

(b) Labeled Faces In the Wild

(c) FERET

Fig. 4. Example Images from the (a) NIST SD 18, (b) Labeled Faces in the Wild, and (c) FERET datasets. Black rectangles added to eye regions of images in (a) in respect of the privacy of the individuals.

Bias in experimental datasets is not specific to research in assessing criminality from face. Another instance of the problem was recently recognized in "kinship detection" research. The kinship problem is to analyze two face images and detect if the persons have a relation such as parent-child or sibling. The KinFaceW-II dataset was assembled and distributed to support research in this area, and has 250 pairs of images for each of Father-Son (F-S), Father-Daughter (F-D), Mother- Son (M-S) and Mother-Daughter. The dataset has been used in research publications by various researchers. But Lopez et al. [18] pointed out that, for each kinship pair, the two face images have been cropped from the same original image, and that two face images cropped from the same larger image share similarity that has nothing to do with the faces. To make the point, they presented results comparing image pairs purely on the chrominance distance between images (no facial analysis at all) showing that chrominance distance actually scored higher in kinship detection than a number of published algorithms. Based on these results, they state, "we strongly recommend that these data sets are no longer used in kinship verification research". Dataset bias in this area led to apparently impressive accuracy that has nothing to do with the phenomenon of interest, and similar bias can readily account for the results in criminality from face research.

A relatively simple principle for evaluating experimental results in facial analytics research is suggested as a reasonable safeguard: no paper that claims to use machine learning to predict labels for the content of images, in this case Criminal and Non-Criminal for face images, should be accepted for publication if its experimental data for the different labels is 100% correlated to different sources. This would capture kinship results in which all true kin image pairs are cropped from a single source image, and criminality from face research in which all Criminal images come from one source and all Non-Criminal images come from a different source.

Some researchers may not accept our argument that pursuit of a criminality-from-face algorithm is doomed from first principles to failure. Rhetorically, for them, what would be a more convincing experimental design for their research? Images of pairs of monozygotic twins in which one twin had a criminal record and one did not, might make a more convincing experimental design. Images of pairs of persons, cropped from the same group photo, where one person has a criminal record and the second person, with no criminal record, is selected as the most similar person in the photo might also be compelling. This sort of dataset might be assembled from photos of sports teams, musical groups, political groups, or other sources. Even in this setting, the evaluation could be abused through "data dredging," whereby training strategies are constantly changed to maximize accuracy over the test set [44]. This could be addressed through the use of a sequestered evaluation, where algorithm submissions are limited.

Finally, there is a question of algorithmic transparency in this setting. If an algorithm designed to detect criminality from face images could be interrogated about how it reached its conclusions, what would it say? If what we have outlined is true, the results would be non-sensical, ideally leading the creator to realize the flaw in their work. Commonly used machine learning technologies are opaque, but interest in the research area of explainable AI, and how it relates to established ethical frameworks, is beginning to change that [45]. By adding internal reasoning about fairness into an algorithm by design, consistent bad decision-making could be detected [46].

## V. Confusion with Models of "First Impressions"

Wu and Zhang cite the work of Princeton psychologist Alexander Todorov as a justification for the plausibility of modeling criminality from faces [35], [32], [34]. However, this justification is based on a mistaken assumption that by modeling the first impressions of subjects viewing a face as Todorov does, one can discern something true about the underlying personality traits for that face's identity. Todorov's research is limited to the social perception of faces, and models his laboratory has published make predictions about what the average person would likely say about a particular face image [24], [31]. These predictions represent a consensus of sorts for various attribute judgements (e.g., trustworthiness, dominance). In Wu and Zhang's words, the existence of consensus judgments for certain attributes allows them to explore the "diagnostic merit of face-induced inferences on an individual's social attributes" [40]. In other words, the extent to which physiognomic cues predict personality traits, which they believe Todorov's work hints at.

But the existence of a consensus attribute judgement for a particular person's appearance does not mean that it holds any truth about their personality. Much to the contrary of Wu and Zhang's claims, Todorov writes in his preface to the book Face Value: the Irresistible Influence of First Impressions [30] that "Psychologists in the early twentieth century found little evidence for the accuracy of first impressions, but the past decade has seen a resurgence of physiognomic claims in scientific journals. We are told that it is possible to discern a person's political leanings, religious affiliation, sexual orientation, and even criminal inclinations from images of their face...A closer look at the modern studies shows that the claims of the new physiognomy are almost as exaggerated as those in the eighteenth and nineteenth centuries." Given the work Todorov has published within social psychology, it is not surprising to learn that he is sharply critical of the idea that one can determine criminality solely by looking at faces.

Wu and Zhang are not the only researchers that have considered Todorov's work in the context of predicting criminality from faces. Valla et al. conducted behavioral studies on the accuracy of people for this task, remarkably finding that groups of subjects were able to discriminate between criminals and non-criminals in some cases [37]. (Wu and Zhang confirmed to us that this is the Cornell University study that they reference in their response to the critiques levied against their paper [41].) Similar to Wu and Zhang, Valla et al. also believe that Todorov's work on first impressions demonstrates a link between social behavior and innate traits. In order to argue this point, they allege that Todorov has a tendency to "shy away from the possibility of accurate impressions" based on physiognomic cues out of "concern that it harkens back to the stigmas associated with social Darwinism." Thus, according to Valla et al., Todorov's findings can be used as a justification for criminality-from-face studies — he simply isn't drawing a strong enough conclusion from his data. Of course, this line of argumentation only makes sense if first impressions can be shown to be reliable predictors of innate behavioral traits. As for Valla et al.'s remarkable finding — they combined mug shots of arrested people with photographs of innocent students on campus in their study, such that the task performed by the subjects was really just dataset discrimination [33].

A more recent study on the convictability of faces, also from Cornell, does make use of photos from the same source to show low, but above chance, accuracy for human subjects on this task [23]. But it ultimately concludes that non-face context is likely a significant driver of decisions, and warns off using the results in a criminal justice context other than attempting to understand how faces are viewed in a social context (in the manner of Todorov).

The sound work that has been done on how humans form subjective first impressions from a face image does not imply that the first impression is actually true. The work that has explored whether humans can accurately determine Criminal / Non-Criminal from a face image runs into the same dataset bias pitfall as work on automated facial analytics for predicting Criminal / Non-Criminal. A persuasive experiment for the alleged phenomenon of humans being able to accurately perceive the criminality of persons from their face image has yet to emerge.

## VI. THE LEGACY OF THE POSITIVIST SCHOOL OF CRIMINOLOGY

Society at large should be very concerned if physiognomy makes a serious resurgence through computer vision. Algorithms that attempt to determine criminality from face images reinforce mistaken beliefs about biology and add a deceitful instrument to the ever-growing digital surveillance toolkit. Such technology is rooted in the Positivist School of criminology, which has long argued that criminals are born, not made [16]. Indeed, Hashemi and Hall directly acknowledge Cesare Lombroso, the 19th century sociologist who founded the Positivist School, as the motivation for their work [14]. Inspired by the newly introduced theories of Charles Darwin[1], Lombroso popularized the technique of facial measurement for predicting criminal tendencies, arguing that heritable flaws manifested themselves in both anatomy and behavior [17]. Lombroso's research was eventually discredited on the grounds that it did not make use of valid control groups [39], but Positivist notions persist in contemporary thinking about criminality [7, 47].

And it is not difficult to understand why this idea is still attractive. Since the dawn of the genetic revolution in biology, the general public has developed a commonly held belief that genes code for complex behaviors (think of the expression "it's in my genes"). Thus it is not a stretch to imagine criminal behaviors having some genetic basis under this regime. But such a simplistic belief is problematic in that it skips several levels of abstraction, ignoring the essential role of learning in human development [15], as well as the interplay between the environment and a nervous system defined by a genetic profile [27]. While there may be some correlation between genes and complex behaviors, the mechanisms are not currently understood, and no evidence of a direct genetic link to criminal behavior exists [21]. A further confound surfaces when behavioral traits must be coupled with some physical manifestation to diagnose criminality. To justify the plausibility of this, one could point to conditions such as the fetal alcohol spectrum disorders, which present with abnormal facial features and anti-social behavioral traits [26]. But the vast majority of criminals in any country do not suffer from such syndromes [20]. Given the variety of mental disorders that do not present with any obvious physical abnormality (e.g., mood disorders, schizophrenia), there can be no expectation that a physical marker associated with criminality will be present even in cases where there is some indirect genetic basis to the behavior that led to a crime.

## VII. SOCIAL IMPLICATIONS OF THIS TECHNOLOGY

The aforementioned misunderstandings about biology have a problematic social implication when Positivist ideas are coupled to systems of mass surveillance. Contemporary theories of criminal control are wrapped in scientific language in order to gain legitimacy within the academy and dodge scrutiny from policy makers [8]. Thus it is convenient to talk about the biology of criminality when one needs to justify the use of a controversial technology. As we have already pointed out, there is a logical disconnect between legal definitions of criminality and the body. Nonetheless, the rise of the surveillance state in the 20th century and surveillance capitalism in the 21st was predicated on the distortion of scientific findings. Here we discuss three particularly troubling scenarios where artificial intelligence (AI) has already been used in a manner that erodes human rights and social trust, which could be further exacerbated by the deployment of ineffective criminality from face algorithms.

The first scenario is the nation-scale use of this technology by a government that mistakenly believes that it works as advertised. There is growing interest in facial analytics for surveillance purposes, and algorithms that assess visual facial attributes have been added to that repertoire [1]. The existing technologies that do actually work have already proven to be controversial. In 2019, marketing material for a smart camera system with an automatic ethnicity detector from the Chinese technology company Hikvision surfaced [22]. In particular, this product was advertised as being able to tell the difference between Han Chinese, the ethnic majority in China, and Uyghurs, an ethnic minority involved in a long-standing conflict with the central government in Beijing. Because Uyghurs do indeed look different than Han Chinese, they can be detected and tracked via automated means. Under similar reasoning, if the same is true of criminals and innocents, then profiling with facial analytics is also possible in that case. Both of these technologies discriminate against an out-group, and, regardless of the correctness of their output, can lead to innocent people being discriminated against at best, and a senseless loss of life at worst.

Related to the first scenario is the second, which is the use of this technology in data driven predictive policing. Instead of widespread deployment, cameras equipped to detect criminals could be installed in more localized "hot spots" to study their movements so that the police would know where to look for criminal activity in the future. There is already a lucrative market for law enforcement products of this nature [11]. While the ability to monitor the activities of potential lawbreakers is tantalizing, problematic racial biases have been found in facial recognition technologies that match surveillance photos to mugshots [12]. Those same racial biases are likely to become manifest in any machine learning-based system using that data, given that they are an artifact of the data itself. It is not hard to imagine criminality-from-face algorithms being trained with the same databases that are currently used for other predictive policing applications. Thus, the best these algorithms can do is reproduce available biases as their decisions, leading to a strongly misleading picture of the criminal presence in an area.

We also find similar problems in the commercial world. One example is the application of personality attribute prediction for job candidate assessment. Machine learning-based personality profiling is now being used as a first-round screening process at some companies [2]. Given the uptick in interest in AI automation, this practice will only spread. In particular, one would expect to see such technology being deployed extensively in the service industry to reduce hiring costs. With more concern in service-oriented businesses about the risk of criminal behavior on the job, there will inevitably be interest in a capability to predict criminality from face. As with the government surveillance and predictive policing scenarios, the risk of this directly leading to discriminatory practices is high.

---

[1] Note that other related movements under the influence of Darwin such as Eugenics (popularized by Francis Galton [42]) and Biological Determinism (more recently promoted by E. O. Wilson [43]) came to conclusions similar to Positivism. Given our focus on criminality from face, we will only discuss the Positivist school, which explicitly treats this topic, in this section.

## VIII. CONCLUSIONS

In spite of the assumption that criminality-from-face is similar to other facial analytics, there is no coherent definition on which to base development of an algorithm. Seemingly promising experimental results in criminality-from-face are easily accounted for by simple dataset bias. The concept that a criminality-from-face algorithm can exist is an illusion, and belief in the illusion is dangerous.

The most innocuous danger of the criminality-from-face illusion is that good researchers will waste effort that could otherwise create solutions that truly would benefit humanity. A larger danger is that government and industry will believe the illusion and expend precious resources on an effort that cannot succeed. The most ominous danger is that belief in the illusion will result in applications being fielded that arbitrarily sort human beings that "*fit the description*" into the categories Criminal and Non-Criminal — with potentially grievous consequences.

## REFERENCES

[1] FLIR Neuro technology: Automate complex decisions faster with deep learning. https://www.flir.com/discover/iis/machine-vision/flir-neuro-technology-automate-complex-decisions-faster-with-deep-learning/.
[2] L. Burke. Your interview with AI. *Inside Higher Ed*, November 4th, 2019.
[3] M. Chen, J. Fridrich, M. Goljan, and J. Lukas. Determining image origin and integrity using sensor noise. *IEEE Transactions on Information Forensics and Security* 3(1):74–90, 2008.
[4] D. Cozzolino and L. Verdoliva. Noiseprint: A CNN-based camera model fingerprint. *IEEE Transactions on Information Forensics and Security* 15:144–159, 2020.
[5] A. Dantcheva, P. Elia, and A. Ross. What else does your biometric data reveal? A survey on soft biometrics. *IEEE Transactions on Information Forensics and Security* 11(3), 2016.
[6] E. Eidinger, R. Enbar, and T. Hassner. Age and gender estimation of unfiltered faces. *IEEE Transactions on Information Forensics and Security* 9(12), 2014.
[7] S. Ferracuti. Cesare Lombroso (1835–1907). *Journal of Forensic Psychiatry*, 7(1):130–149, 1996.
[8] M. Foucault. Discipline and Punish: The Birth of the Prison. Vintage, 2012.
[9] S. Fu, H. He, and Z.-G. Hou. Learning race from face: A survey. *IEEE Transactions on Pattern Analysis and Machine Intelligence* 36(12):2483–2509, 2014.
[10] Y. Fu, G. Guo, and T.S. Huang. Age synthesis and estimation via faces: A survey. *IEEE Transactions on Pattern Analysis and Machine Intelligence* 32(11), 2010.
[11] Globe Newswire. Police modernization market to reach $59.9B by 2025. https://www.prnewswire.com/news-releases/law-enforcement–police-modernization-market–2020-2025-300980287.html, 2019.
[12] A. Harmon. As cameras track Detroit's residents, a debate ensues over racial bias. *The New York Times*, July 8, 2019. https://www.nytimes.com/2019/07/08/us/detroit-facial-recognition-cameras.html, 2019.
[13] Harrisburg University. Research brief on facial recognition software. https://harrisburgu.edu/hu-facial-recognition-software-identifies-potential-criminals/, 2020.
[14] M. Hashemi and M. Hall. Criminal tendency detection from facial images and the gender bias effect. *Journal of Big Data*, 7(2), 2020.
[15] B. M. Lake, T. D. Ullman, J. B. Tenenbaum, and S. J. Gershman. Building machines that learn and think like people. *Behavioral and Brain Sciences*, 40, 2017.
[16] J. Law and E. A. Martin. A Dictionary of Law. OUP Oxford, 2014.
[17] G. Lombroso. Criminal man, according to the classification of Cesare Lombroso. Good Press, 2019.
[18] M. B. Lopez, E. Boutellaa, and A. Hadid. Comments on the "kinship face in the wild" data sets. *IEEE Transactions on Pattern Analysis and Machine Intelligence* 11(38):2342–2344, 2016.
[19] NIST. NIST Special Database 18 - NIST Mugshot Identification Database (MID).
[20] S. Popova, S. Lange, D. Bekmuradov, A. Mihic, and J. Rehm. Fetal alcohol spectrum disorder prevalence estimates in correctional systems: a systematic literature review. *Canadian Journal of Public Health*, 102(5):336–340, 2011.
[21] A. Raine. The psychopathology of crime: Criminal behavior as a clinical disorder. Elsevier, 2013.
[22] [C. Rollet. Hikvision markets Uyghur ethnicity analytics, now covers up. https://ipvm.com/reports/hikvision-uyghur, 2019.
[23] C. E. Royer. Convictable faces: Attributions of future criminality from facial appearance. 2018.
[24] C. P. Said and A. Todorov. A statistical model of facial attractiveness. *Psychological Science*, 22(9):1183–1190, 2011.
[25] E. Sariyanidi, H. Gunes, and A. Cavallaro. Automatic analysis of facial affect: A survey of registration, representation, and recognition. *IEEE Transactions on Pattern Analysis and Machine Intelligence* 37(6):1113–1133, 2015.
[26] R.J. Sokol, V. Delaney-Black, and B. Nordstrom. Fetal alcohol spectrum disorder. *JAMA*, 290(22):2996–2999, 2003.
[27] E.S. Spelke. Nativism, empiricism, and the origins of knowledge. *Infant Behavior and Development*, 21(2):181–200, 1998.
[28] J. D. Stein. How to make an innocent client plead guilty. *Washington Post*, 12 January 2018. https://www.washingtonpost.com/opinions/why-innocent-people-plead-guilty/2018/01/12/e05d262c-b805-11e7-a908-a3470754bbb9 story.html.
[29] J. Thevenot, M. B. López, and A. Hadid. A survey on computer vision for assistive medical diagnosis from faces. *IEEE Journal of Biomedical and Health Informatics,* 22(5):1497–1511, 2018.
[30] A. Todorov. Face value: The irresistible influence of first impressions. Princeton University Press, 2017.
[31] A. Todorov, R. Dotsch, J.M. Porter, N.N. Oosterhof, and V.B. Falvello. Validation of data-driven computational models of social perception of faces. *Emotion*, 13(4):724, 2013.
[32] A. Todorov, V. Loehr, and N. N. Oosterhof. The obligatory nature of holistic processing of faces in social judgments. *Perception*, 39(4):514–532, 2010.
[33] A. Todorov, C. Y. Olivola, R. Dotsch, and P. Mende-Siedlecki. Social attributions from faces: Determinants, consequences, accuracy, and functional significance. *Annual Review of Psychology*, 66:519–545, 2015.
[34] A. Todorov and N. N. Oosterhof. Modeling social perception of faces. *IEEE Signal Processing Magazine*, 28(2):117–122, 2011.
[35] A. Todorov, M. Pakrashi, and N. N. Oosterhof. Evaluating faces on trustworthiness after minimal time exposure. Social *Cognition*, 27(6):813–833, 2009.
[36] A. Torralba and A. A. Efros. An unbiased look at dataset bias. In IEEE CVPR, 2011.
[37] J. M. Valla, S. J. Ceci, and W. M. Williams. The accuracy of inferences about criminality based on facial appearance. *Journal of Social, Evolutionary, and Cultural Psychology*, 5(1):66, 2011.
[38] C. Watson and P. Flanagan. NIST Special Database 18 – Mugshot Identification Database. 2016.
[39] M. E. Wolfgang. Pioneers in criminology: Cesare Lombroso (1825-1909). *J. Crim. L. Criminology & Police Sci.*, 52:361, 1961.
[40] X. Wu and X. Zhang. Automated inference on criminality using face images. https://arxiv.org/abs/1611.04135, 2016.
[41] X. Wu and X. Zhang. Responses to critiques on machine learning of criminality perceptions (addendum of arxiv: 1611.04135). arXiv preprint arXiv:1611.04135, 2016.

[42] M. Brookes. Extreme Measures: The Dark Visions and Bright Ideas of Francis Galton. Bloomsbury Pub. Ltd., 2004.
[43] E. O. Wilson. Sociobiology: The New Synthesis, Twenty-Fifth Anniversary Edition. Belknap Press, 2000.
[44] M. Kearns and A. Roth. The Ethical Algorithm: The Science of Socially Aware Algorithm Design. Oxford University Press, 2019.
[45] V. Dignum. Responsible Autonomy. In IJCAI, 2017.
[46] M. Kusner, J. Loftus, C. Russell, and R. Silva. Counterfactual Fairness. In NeurIPS, 2017.
[47] D. G. Horn. The Criminal Body: Lombroso and the Anatomy of Deviance. Routledge, 2003.
[48] Gary B. Huang, Manu Ramesh, Tamara Berg, and Erik Learned-Miller. Labeled Faces in the Wild: A Database for Studying Face Recognition in Unconstrained Environments. University of Massachusetts, Amherst, Technical Report 07-49, October, 2007
[49] Phillips, P. J., Wechsler, H., Huang, J., & Rauss, P. J. (1998). The FERET database and evaluation procedure for face-recognition algorithms. *Image and Vision Computing*, 16(5), 295-306.
[50] The Innocence Project. https://www.innocenceproject.org/cases/thomas-doswell/)

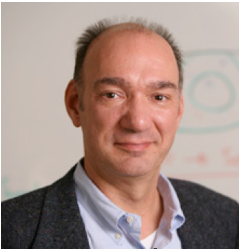

**Kevin W. Bowyer** (Fellow, IEEE) received the Ph.D. degree in computer science from Duke University, Durham, NC, USA.

He is the Schubmehl-Prein Family Professor of computer science and engineering with the University of Notre Dame, Notre Dame, IN, USA.

Prof. Bowyer received the Technical Achievement Award from the IEEE Computer Society, with the citation "for pioneering contributions to the science and engineering of biometrics." He served as the Editor-in-Chief for the IEEE TRANSACTIONS ON PATTERN ANALYSIS AND MACHINE INTELLIGENCE. He is currently serving as the Editor-in-Chief for the IEEE TRANSACTIONS ON BIOMETRICS, BEHAVIOR AND IDENTITY SCIENCE. In 2019, he was elected as a fellow of the American Association for the Advancement of Science (AAAS).

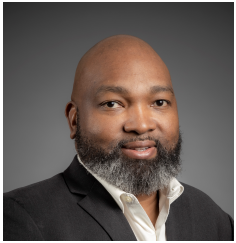

**Michael C. King** (Member, IEEE) received the Ph.D. degree in electrical engineering from North Carolina Agricultural and Technical State University, Greensboro, NC, USA.

He joined the Harris Institute for Assured Information, Florida Institute of Technology, Melbourne, FL, USA, as a Research Scientist in 2015 and holds a joint appointment as an Associate Professor of computer engineering and sciences. Prior to joining academia, he served for more than ten years as a Scientific Research/Program Management Professional with the United States Intelligence Community. While in government, he created, directed, and managed research portfolios covering a broad range of topics related to biometrics and identity to include: advanced exploitation algorithm development, advanced sensors and acquisition systems, and computational imaging. He crafted and led the Intelligence Advanced Research Projects Activity's Biometric Exploitation Science and Technology Program to transition technology deliverables successfully to several Government organizations.

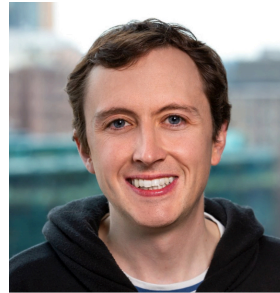

**Walter J. Scheirer** (Senior Member) received the MS degree in computer science from Lehigh University, in 2006, and the Ph.D. degree in engineering from the University of Colorado, Colorado Springs, CO, USA, in 2009.

He is an Associate Professor with the Department of Computer Science and Engineering, University of Notre Dame. Prior to joining the University of Notre Dame, he was a Postdoctoral Fellow with Harvard University from 2012 to 2015, and the Director of Research and Development with Securics, Inc., from 2007 to 2012. His research interests include computer vision, machine learning, biometrics, and digital humanities).

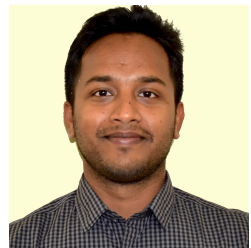

**Kushal Vangara** is a Ph.D. student at the Florida Institute of Technology. He received the B. Tech. degree in Electronics and Communications Engineering from Jawaharlal Nehru Technological University, India, in 2014 and the M.S. degree in Information Assurance and Cybersecurity from the Florida Institute of Technology, in 2018. He is currently pursuing the Ph.D. degree in computer science at Florida Tech. His research interests include biometrics, computer vision, data science and deep learning.